\documentclass[runningheads]{llncs}
\usepackage{graphicx}
\usepackage{rotating}

\usepackage[cmex10]{amsmath}
\usepackage{amssymb}

\begin{document}
\title{New feature for Complex Network based on Ant Colony Optimization}
\titlerunning{New feature for Complex Network based\\
on Ant Colony Optimization}
%\orcidID{0000-0003-2942-0853}
\author{Josimar E. Chire-Saire\inst{1} }%\and Roger Temam\inst{2}
%Jeffrey Dean \and David Grove \and Craig Chambers \and Kim~B.~Bruce \and
%Elsa Bertino}
%
\authorrunning{Chire Saire, J.} % abbreviated author list (for running head)
% \tocauthor{Ivar Ekeland, Roger Temam, Jeffrey Dean, David Grove,
% Craig Chambers, Kim B. Bruce, and Elisa Bertino}
%
\institute{Institute of Mathematics and Computer Science (ICMC),\\  University of São Paulo (USP), São Carlos, SP, Brazil \\
\email{jecs89@usp.br}} 

\maketitle              % typeset the header of the contribution
\begin{abstract}
Low level classification extracts features from the elements, i.e. physical to use them to train a model for a later classification. High level classification uses high level features, the existent patterns, relationship between the data and combines low and high level features for classification. High Level features can be got from Complex Network created over the data. Local and global features are used to describe the structure of a Complex Network, i.e. Average Neighbor Degree, Average Clustering.% These features represents the architecture of the Network to use it for High Classification Problem.
The present work proposed a novel feature to describe the architecture of the Network following a Ant Colony System approach. The experiments shows the advantage of using this feature because the sensibility with data of different classes.

\end{abstract}

\section{Introduction}

The nature has a special order to establish how the things work, human beings want to understand how nature works and from many centuries they have studied the environment from Math, Physics, Chemistry, Biology and other fields. Many of the studies show the existence of many systems, repeated structures(pattern) are present in the form how they interconnect each other, i.e. cell systems, solar system, electric system and others.

Then, according to the object of study, the presence of networks can be high or low. From Social fields, the study of the interaction of people in one specific city, state or country involve a network of many kind of relationships. From Biological fields, many cells, proteins interact each other following some specific behaviour, and networks are involved again. Physics has started to study many physical phenomenons using Networks.

Complex Network(CN) is a field which used graphs to represent entities and any kind of relationships present between them. A graph is represented for nodes and edges, then the nodes can represent objects and the edges, the relationships, i.e. social interaction, chemical reaction, distance, etc. Many measurements can be extracted to represent internal, external relationships.

Ant Colony Optimization\cite{col1991} is a bio-inspired algorithm based on ants behaviour, how they find the good path to get the food. These kind of algorithms are proposed to solve optimization problems and find a good solution in a reasonable computational time.

The present paper introduce a novel feature to explore the behaviour of Complex Networks based on Minimal Spanning Tree, Travelling Sale Problem. The objective is to find the minimal path that connects all the nodes, this process can be expensive but the process can be performed in a reasonable time with Ant Colony algorithm and use the path as measurement of the Complex Network. 

This paper is organized as follows: section 2 describes a literature review on Complex Network features, section 3 describes the proposal, section 4 describes the experiments performed to test the proposal, and finally in section 5, some conclusions are presented.

\section{LITERATURE'S REVIEW}
\label{sec:2}

Traditional data classification algorithms consider only low level features(physical features) to build the model but human brain has low/high level learning. The work of Silva\cite{Silva2011} proposes the construction of a network over the features to find the existing patterns(high level features), then the proposal is tested on handwritten digits recognition using the next equation for the classification: 
\begin{equation}
M_y^c = (1 - \lambda)C_y^c + \lambda H_y^c
\label{eq:lhequation}
\end{equation}, where $M_y^c$ represents the combined classification(low, high level algorithms) when evaluating an instance $y$ for the class $c$. Also, C and H are the association between instance $y$ for the class $c$ for low and high level features and $\lambda$ $\in$ [0,1], which is an user-controllable variable.
The results shows a meaningful improvement rate, demonstrating the previous hypothesis of mixing low and high level features. An extension of the proposal is presented by Silva\cite{Silva2012} with many artificial datasets with visual patterns for classification tasks showing the high importance of high level features when the complexity increases, besides a test with many well-know datasets like: iris, wine, glass and another showing how the proposal of using high level features can improve the performance of other well-know classifiers as Bayesian Networks, Weighted kNN, Fuzzy C4.5, Multilayer Perceptron and Fuzzy M-SVM. The two previous papers used assortativity, clustering coefficient, average degree for high level features but Silva in his next work\cite{Silva2013} proposes a new high level feature to describe the network using tourist walks, a weighted combination of cycle and transient lengths. The experiments are performed over three scenarios: pure, networkless and network, with the same datasets: iris, wine, etc. and the classification algorithms: Bayesian Networks, Weighted k-NN, Multilayer Perceptron and Fuzzy M-SVM. And the final work of Silva\cite{Silva20131} performs an analysis of the proposal based on tourist walks over hand-written recognition. 

At the same time, Neto\cite{Neto2103} made the proposal of using the entropy of the network to analyze the belonging of one element to the network. The proposal was analyzed over iris dataset, seeds dataset and compare with Multilayer Perceptron, SVM using one artificial dataset with a visual square pattern showing a better performance.

All the previous papers were using a combined classification using low/high level features but Carneiro\cite{Carneiro2013} proposes a high level model where low level is embedded and uses a new complex network measure named component efficency ,and compares with Decision Tree and Support Vector Machine using the next datasets: iris, ecoli, opt. digits, SpectFHeart with good results. After, Carneiro\cite{Carneiro2016} proposes a work using SL-PSO(Social Learning Particle Swarm Optimization) to find the best combination of parameters($\lambda$) for the hybrid model proposed using a quality function.

Later, Colliri\cite{Colliri2018} presented a model based only in high level features, characteristics from the network. The work uses the complex network measurements(average degree, assortativity, average local clustering coefficient, transitivity, average shortest path, second moment of degree distribution) to analyze the belonging of the elements to the network and the impact over the network. Eight artificial datasets, nine well-know datasets are the set for testing, the results shows the promissory way for the proposal. 

Meanwhile, Carneiro\cite{Carneiro2017} was working on an application of high level classification(low/high level features) for Sematinc Role Labeling(SRL), this task involves the identification and classification of the argument in one and indicates the semantic relationship between an event and the participants. The experiment used PropBank-br(Brazilian Portuguese Corpus), this corpus present scarcity of annotated data and very imbalanced distributions. The results revealed the power of using the proposal for this task.

\section{PROPOSAL}

The proposal starts with the definition of a Complex Network using, there is a Graph $G$ with nodes $V$ and edges $E$ and the existent relationships between nodes are represented by edges.
Then if the Complex Network is a graph, common graph measurements can be helpful like to find the shortest path that connects all the nodes. Then, the Complex Network is a big graph with many relationships and the measurement to analyze the network is the shortest path so the proposal is to find this measurements using Ant Colony Algorithm following the criterion of Colliri\cite{Colliri2018}.

\subsection{Ant Colony Algorithm}

\begin{figure}[h]
  \centerline{
  \includegraphics[width = 0.7\columnwidth]{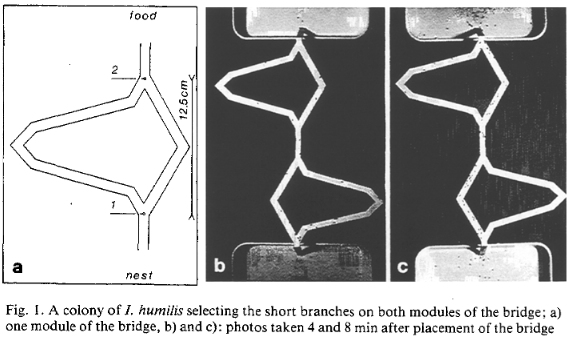}
  }
\caption{ Image\cite{Goss1989} with the results of Goss' experiment }
\label{fig:ant_exp}
\end{figure}

Goss\cite{Goss1989} observed the way how ants search food, he found three main steps: initially randomly each ant search a path, after some time many of the ants follow one unique path, then they have a indirect communication(pheromone) to tell each other which is the best path. One summary of the experiment performed by Goss is presented in Fig. \ref{fig:ant_exp}.

Dorigo\cite{Dorigo1992} realized this behaviour of the ants to find the path for food was similar to Travelling Salesman Problem(TSP) and proposed the next algorithm:
\begin{itemize}
    \item Each artificial ant starts in one node of one graph G
    \item Each ant finds a path(build a solution) following the pheromone deposited in each edge using Eq. \ref{eq:find_sol}
    
    \begin{equation}
        p_{i,j}^k =
        \begin{cases}
        \frac{\tau_{i,j}^\alpha (t) \cdot \eta_{i,j}^\beta}{\sum_{j \in J^k}\tau_{i,j}^\alpha (t) \cdot \eta_{i,j}^\beta} , & \quad if j \in J^k\\
        0 , & \quad otherwise\\
        \end{cases}
        \label{eq:find_sol}
    \end{equation}
    
    where $p_{i,j}^k(t)$ is the probability of ant $k$ follows the edge (i,j) in the iteration $t$, $J^k$ is the list of edges not visited yet, $\tau_{i,j}$ is the quantity of pheromone in the edge (i,j), $\eta_{i,j}$ is the information of quality of this edge(usually the inverse of the distance) and $\alpha,\beta$ are parameters which represent the influence of $\tau,\eta$ respectively.

    \item After the creation of every path, the pheromone is modified according the quality of the generated solution using Eq. \ref{eq:upd_phe1}, \ref{eq:upd_phe2}.
    
    \begin{equation}
        \tau_{i,j} (t+1) = (1- \rho) \cdot \tau_{i,j}(t) + \rho \cdot \delta \tau_{i,j}
    \label{eq:upd_phe1}
    \end{equation}
    
    \begin{equation}
        \delta \tau_{i,j} =
        \begin{cases}
        \frac{1}{f(S)} , & \quad if (i,j) \in S\\
        0 , & \quad otherwise\\
        \end{cases}
    \label{eq:upd_phe2}
    \end{equation}
    
    where $\rho$ is the rate of vaporization of pheromone, $\delta \tau_{i,j}$ is the existent quantity of pheromone in the edge (i,j) and $f(S)$ is the total cost of the solution $S$.
    
\end{itemize}

\subsection{Building the Complex Network}

This subsection explains how to create the complex network from the data:

\begin{itemize}
    \item Use $K$-means to find $K$ cluster and identify the label of each instance
    \item Use the labels to create the complex network, i.e. the instances with label 0 belongs to $CN_0$, instances with label 1 to $CN_1$ and so on.
    \item Each instance of class 0 as node and create an edge between two different instance of the same class and repeat the process for all the classes.
\end{itemize}{}

\subsection{Analysis the change}
The objective is measure how the purposed metric can change when an element of a different class is inserted to one complex network and the opposite.
The analysis is performed using a visualization of the variation of the metric with the insertion of many elements from the same and different class.

\section{EXPERIMENTS AND RESULTS}

The experiments were performed to test the efficiency of the proposal to represent or describe the changes after an insertion of a new element to the initial classes over one artificial dataset and real world datasets. The datasets for the experiments are:

\begin{itemize}
    \item Artificial dataset with circular pattern using a Random Generator Engine with a Normal Distribution($\mu$,$\sigma$), 
    \item Artificial dataset, line structure pattern
    \item Artificial dataset, combining circular and line patterns
    % \item Artificial dataset, combining line and square pattern
    \item Artificial dataset with a polygons structure
    \item Iris Dataset
    \item Wine Dataset
    % \item Breast Cancer Dataset
\end{itemize}

\subsection{Artificial Dataset 1}
The next dataset is of dimensionality two and has the following features:

\begin{itemize}
    \item First class: $mu$ = [2,2], $sigma$ = [0.7,0.7]
    \item Second class: $mu$ = [15,15], $sigma$ = [0.2,0.2]
\end{itemize}

\begin{figure}[h]
\centerline{
\includegraphics[width = 0.45\columnwidth]{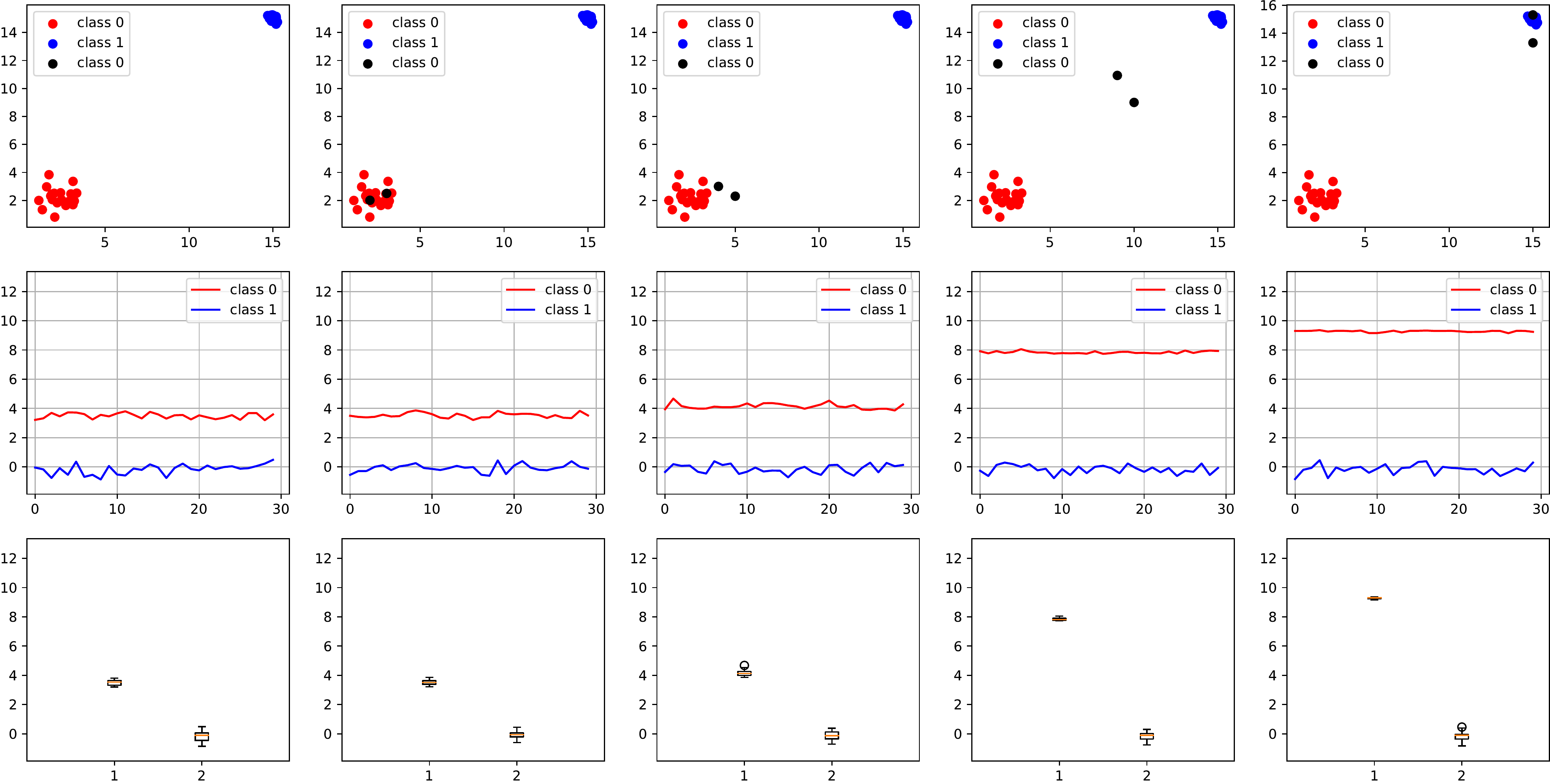}
\includegraphics[width = 0.45\columnwidth]{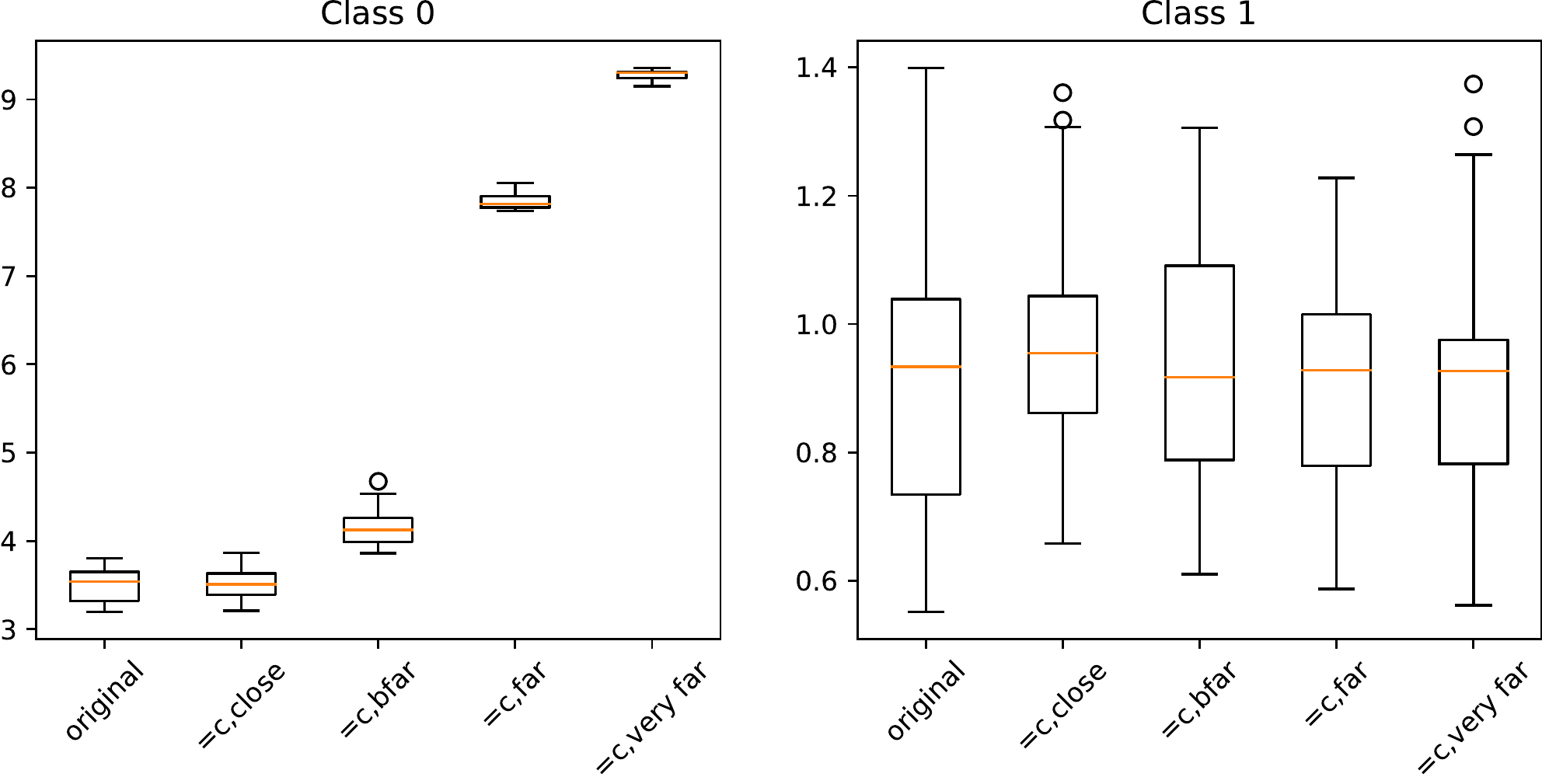}
}
\caption{ Summary of the results, boxplot per class - Dataset 1}
\label{fig:4_art1}
\end{figure}

The results are presented on Fig. \ref{fig:4_art1}. Left side of image, presents the summary of the outcomes.

\begin{itemize}
    \item First, measure the structure in initial state, any change.
    \item Second, add data from the same class(black points), measure the changes.
    \item Third, add points which are closed of the class, measure again.
    \item Next step, add points in an intermediate position.
    \item Finally, add points which belongs to other structure, measure it again.
\end{itemize}

It is possible to check the difference from the third step for class 0(red points), see column 3,4,5 of Fig. \ref{fig:4_art1} and to confirm the differences, right side presents boxplots of each step, then the median is changing from the third phase for class 0. And, the median of class 1 is preserved during the experiments.

\subsection{Artificial Dataset 2}

This dataset is using a line structure, to test the proposal. From this artificial and for next ones, the scenarios are: initial state, adding points to intermediate distance and further or points which belongs to another class.

\begin{figure*}[h]
\centerline{
\includegraphics[width = 0.45\columnwidth]{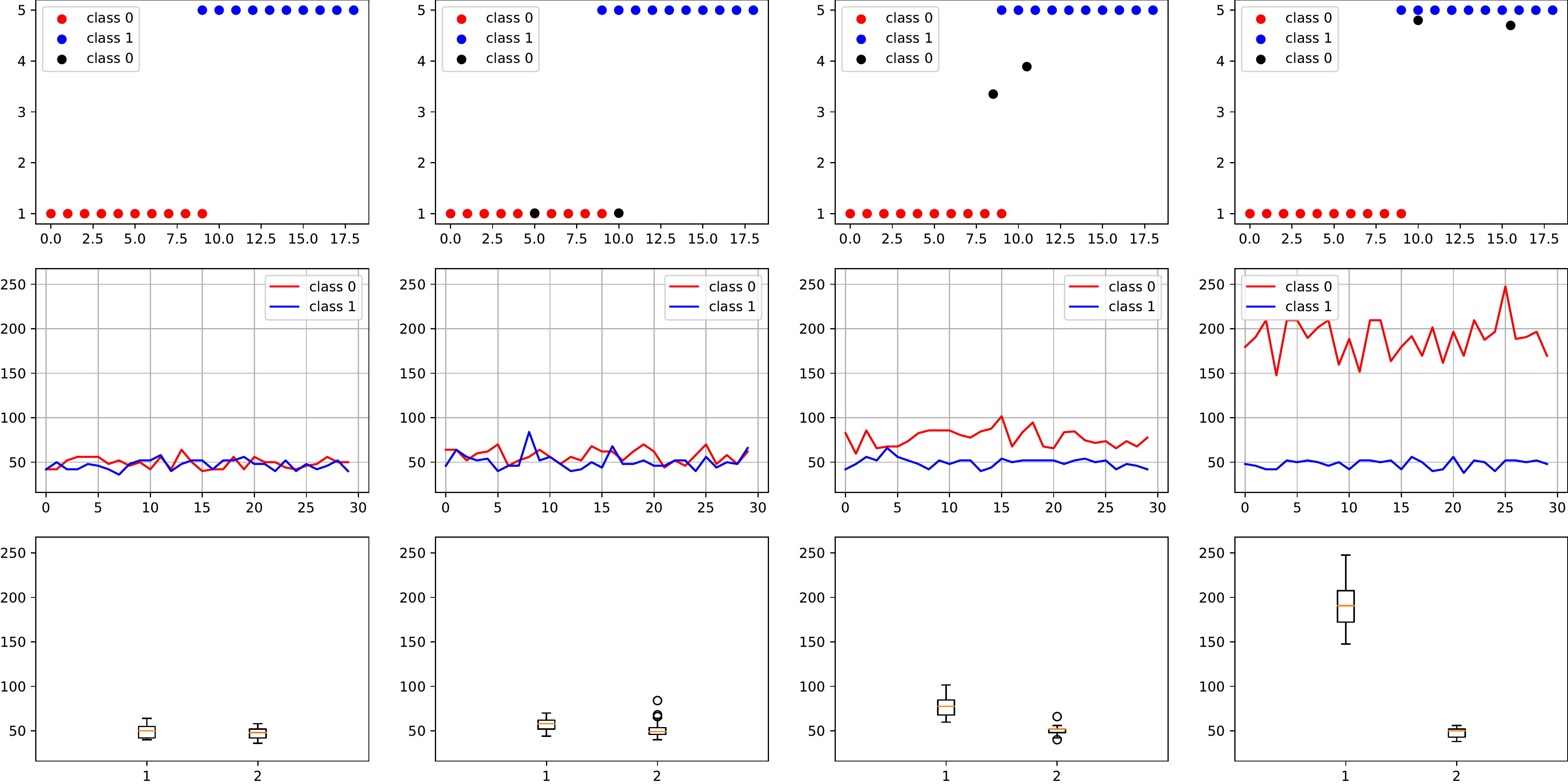}
\includegraphics[width = 0.45\columnwidth]{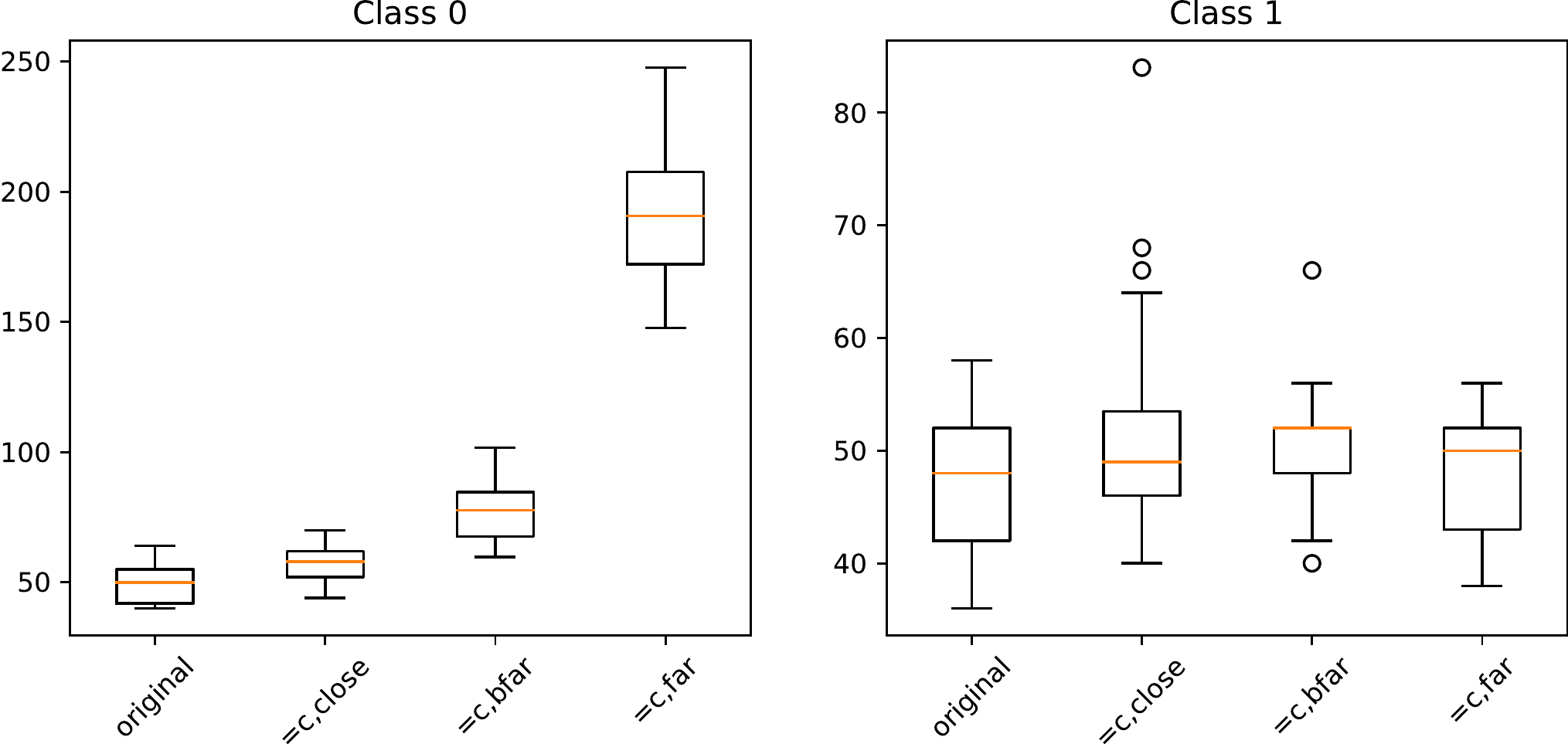}
}
\caption{  Summary of the results, boxplot per class - Dataset 2 }
\label{fig:4_art2}
\end{figure*}

Considering column two, three of the previous image, a modification of the values are evident. Besides, boxplot can show how the median has changed after adding values which not belong to class 0.

\subsection{Artificial Dataset 3}

This set of points is a combination of previous patterns to test the feature base on ACO.

\begin{figure*}[h]
\centerline{
\includegraphics[width = 0.45\columnwidth]{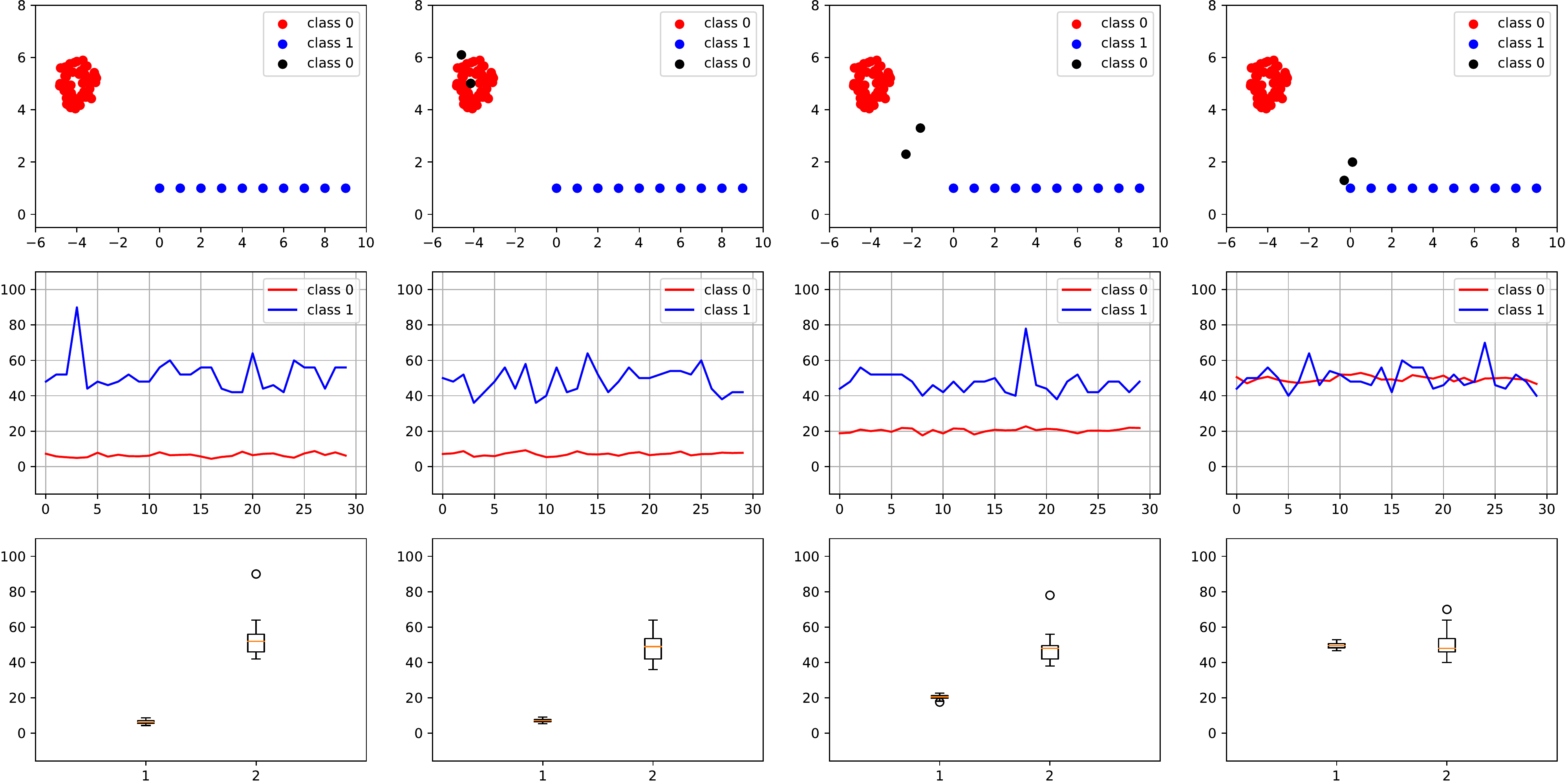}
\includegraphics[width = 0.45\columnwidth]{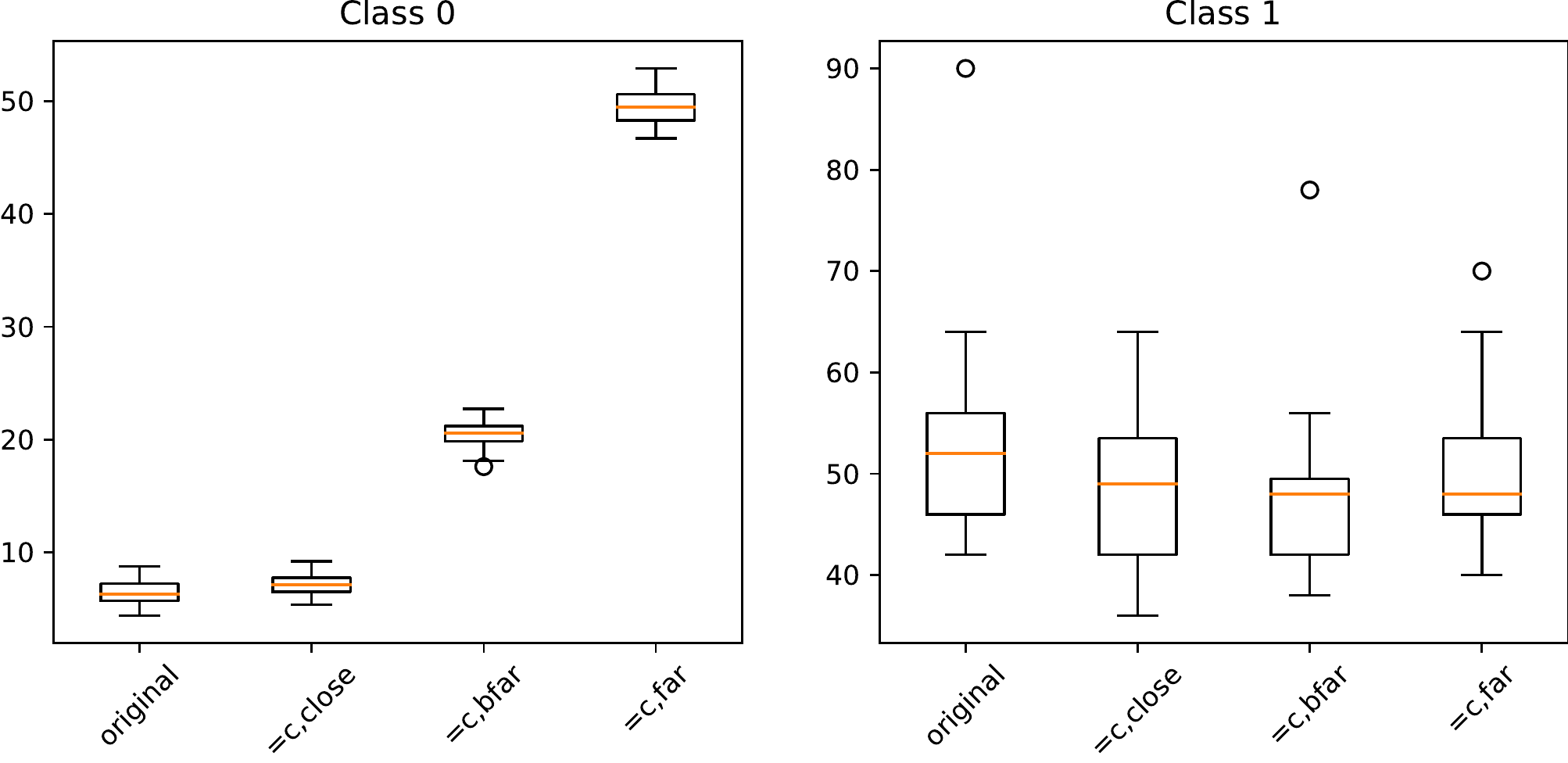}
}
\caption{  Summary of the results, boxplot per class - Dataset 3 }
\label{fig:4_art3}
\end{figure*}

Results are presented in Fig. \ref{fig:4_art3}, the median of values has changed from third phase, showing the sensitivity of the proposal for points of different class. Right side has stable values in the median.

% \subsection{Artificial Dataset 4}

% \begin{figure*}[h]
% \centerline{
% \includegraphics[width = 0.45\columnwidth]{./graphics/exp4_data_value_bp.pdf}
% \includegraphics[width = 0.45\columnwidth]{./graphics/exp4_bp.pdf}
% }
% \caption{  Summary of the results, boxplot per class - Dataset 4 }
% \label{fig:4_art3}
% \end{figure*}

\subsection{Artificial Dataset 4}

\begin{figure}[h]
\centerline{
\includegraphics[width = 0.45\columnwidth]{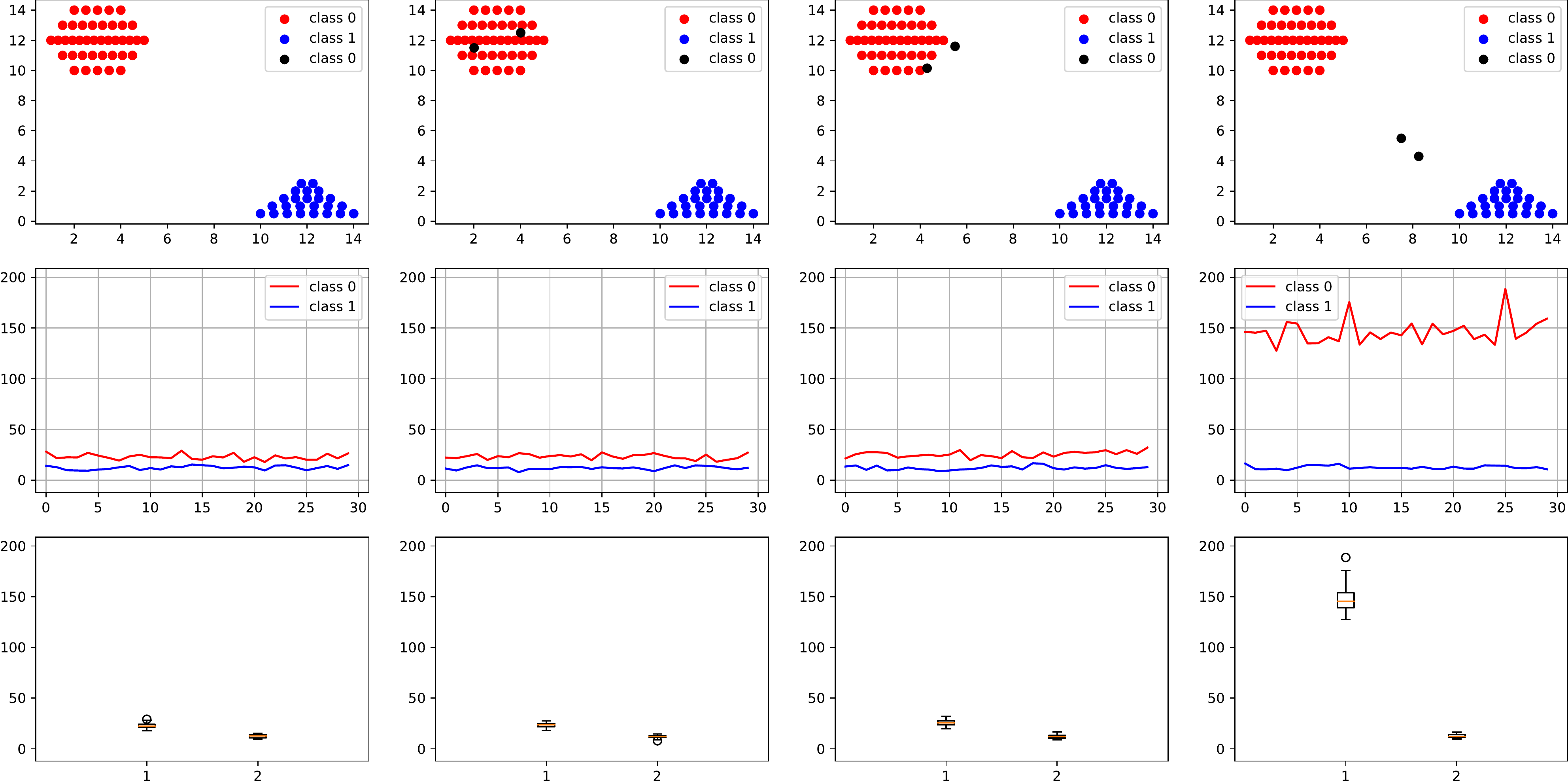}
\includegraphics[width = 0.45\columnwidth]{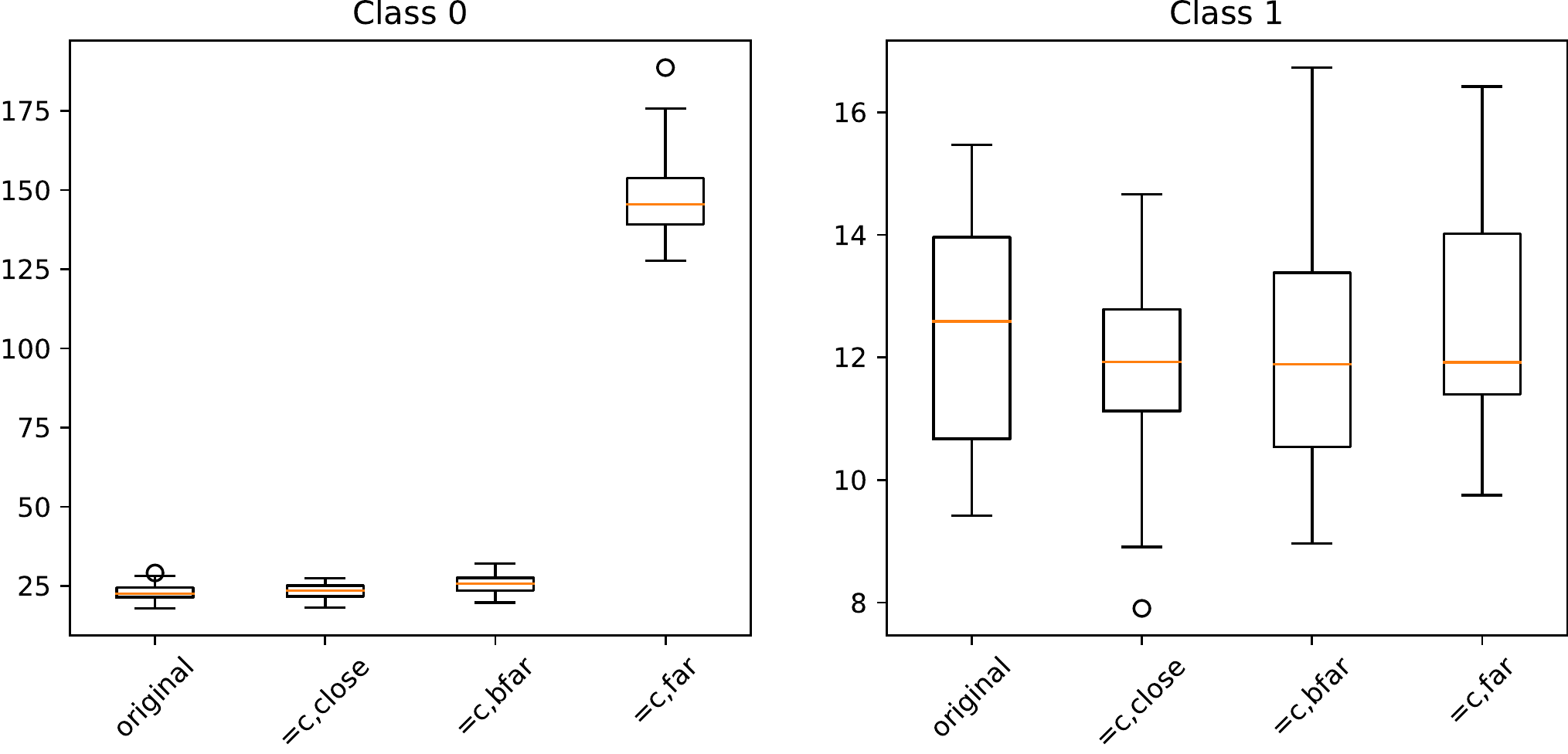}
}
\caption{  Summary of the results, boxplot per class - Dataset 4 }
\label{fig:4_art4}
\end{figure}

This dataset was created considering structure with a polygon form. The graphic \ref{fig:4_art4} show the difference when points which not belong to class are added to the Complex Network.

\subsection{Iris Dataset}

The set of data has three classes(setosa, versicolor, virginica) and 50 instance for each class, each class is the kind of iris plant with four feature.  The experiments are presented on Fig. \ref{fig:4_iridat}:

\begin{figure}[h]
  \centerline{
  \includegraphics[width = 0.3\columnwidth]{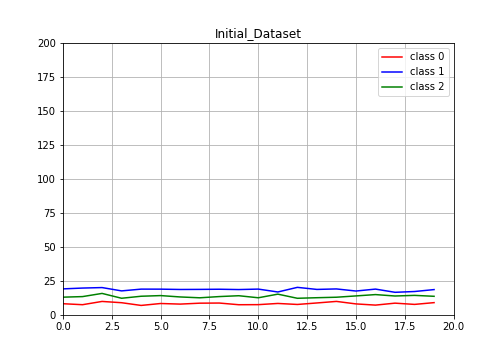}
  \includegraphics[width = 0.3\columnwidth]{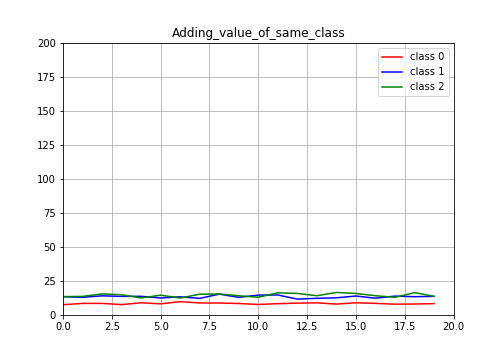}
  \includegraphics[width = 0.3\columnwidth]{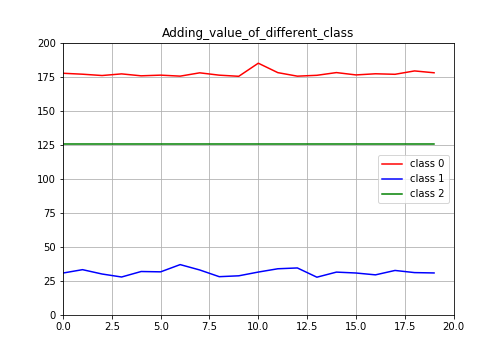}
  }
\caption{ Results with Iris Dataset }
\label{fig:4_iridat}
\end{figure}

First image presents the values in initial state, mid one when points of the same class are added and right one, with addition of points which are not in the class. It possible notice the alteration after adding points of different class to the measurement.

\subsection{Wine Dataset}
The dataset is the result of chemical analysis of wines from many regions of Italy. This datase has 13 features, three classes. The experiments are presented on Fig. \ref{fig:4_windat}:

\begin{figure*}[h]
  \centerline{
  \includegraphics[width = 1.0\columnwidth]{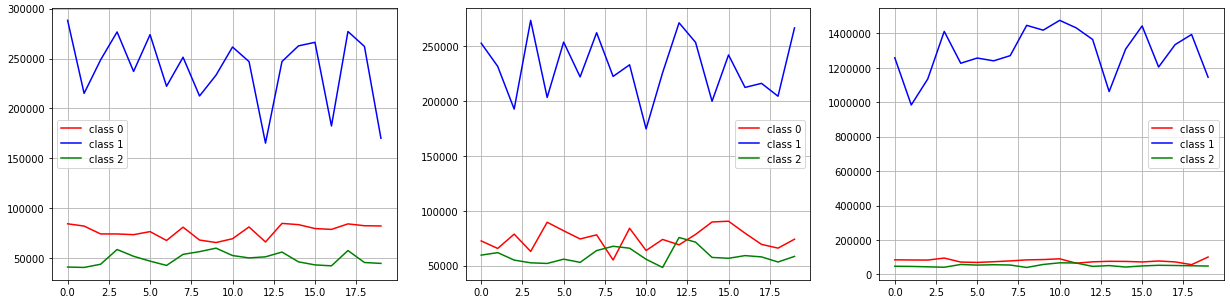}
  }
\caption{ Results with Wine Dataset }
\label{fig:4_windat}
\end{figure*}

Adding values of different class to class 1 creates a big change to the outcome of the measurements, this shows the sensitivity of the proposal to points of different classes.

% % \subsection{Breast Cancer Dataset}

% % The Breast Cancer Wisconsin dataset are the computation from digital image from a fine needle aspirate of a breast mass. The dataset describe 32 characteristics of the cell nuclei present in the image and the class represent benign or malignant tumor. The experiments are presented on Fig. \ref{fig:4_windat}:

% % \begin{figure*}[h]
% %   \centerline{
% %   \includegraphics[width = 2.0\columnwidth]{./graphics/breast.png}
% % %   \includegraphics[width = 0.7\columnwidth]{./graphics/art_}
% % %   \includegraphics[width = 0.7\columnwidth]{./graphics/art_}
% %   }
% % \caption{ Results with Breast Cancer }
% % \label{fig:4_windat}
% % \end{figure*}

% The preliminary results presented in Fig. \ref{fig:4_artdat}, \ref{fig:4_iridat}, \ref{fig:4_windat} shows how sensitive can be the measurement based on ACO to identify if one inserted element belongs or not to the Network(class).

\section{CONCLUSION}

In this paper, we present a new feature to describe the topology of a Complex Network using Ant Colony Optimization to analyze the change before/after insertion of one element of same/different class, the results show the efficiency to represent the change after the insertion of one different element and the preservation after adding one element of the same class. This proposal is open for modifications and improvements for its performance.

\bibliographystyle{splncs04}
\bibliography{biblio.bib}

\end{document}